\documentclass[twoside]{article}
\usepackage[accepted]{aistats2015}
\usepackage{hyperref}
\usepackage{url}

\usepackage[english]{babel}
\usepackage[utf8]{inputenc}

\usepackage{amsmath}
\usepackage{amsfonts}
\usepackage{amssymb}
\usepackage{dsfont}
\usepackage{amsthm}

\usepackage{tikz}
\usetikzlibrary{decorations.pathreplacing}
\usepackage{graphicx}
\usepackage{wrapfig}
\usepackage{color}

\newtheorem{theorem}{Theorem}
\newtheorem{lemma}{Lemma}
\newtheorem{proposition}{Proposition}

\newtheorem{definition}{Definition}

\newtheorem{remark}{Remark}
\newtheorem{algo}{Algorithm}
\def\EM{\textsc{EM}}
\def\mv{\textsc{MV}}
\def\roc{\textsc{ROC}}
\def\S{\mathcal{S}}
\newcommand{\crit}{\mathcal{C}}
\newcommand{\point}{\,\cdot\,}

\date{}

\begin{document}

%

%

\twocolumn[

\aistatstitle{On Anomaly Ranking and Excess-Mass Curves}

\aistatsauthor{ Nicolas Goix \And Anne Sabourin \And Stéphan Clémençon }

\aistatsaddress{ 
UMR LTCI No. 5141 \\
Telecom ParisTech/CNRS \\
Institut Mines-Telecom \\
Paris, 75013, France 
\And UMR LTCI No. 5141 \\
Telecom ParisTech/CNRS \\
Institut Mines-Telecom \\
Paris, 75013, France 
\And UMR LTCI No. 5141 \\
Telecom ParisTech/CNRS \\
Institut Mines-Telecom \\
Paris, 75013, France } 
]

\begin{abstract}
Learning how to rank multivariate unlabeled observations depending on their degree
of abnormality/novelty is a crucial problem in a wide range of
applications. In practice, it generally consists in building a real valued "scoring" function on the feature space
so as to quantify to which extent observations should be considered as abnormal.
 In the 1-d situation, measurements are generally
considered as ”abnormal” when they are remote from central measures
such as the mean or the median. Anomaly detection then relies on tail
analysis of the variable of interest. Extensions to the multivariate setting are far from straightforward and it is precisely the main purpose of this paper to introduce a novel and convenient (functional) criterion for measuring the performance of a scoring function regarding the anomaly ranking task, referred to as the \textit{Excess-Mass} curve ($\EM$
curve). In addition, an adaptive algorithm for building a scoring function based on unlabeled data $X_1,\; \ldots,\; X_n$ with a nearly optimal $\EM$ is proposed and is analyzed from a statistical perspective.
\end{abstract}

\section{Introduction}

In a great variety of applications  (\textit{e.g.} fraud detection, distributed fleet monitoring, system management in data centers), it is of crucial importance to address anomaly/novelty issues from a ranking point of view. In contrast to novelty/anomaly detection (\textit{e.g.} \cite{Kolt97, VertVert, SPSSW01, SHS05}), novelty/anomaly ranking is very poorly documented in the statistical learning literature (see \cite{VCTWMS} for instance). However, when confronted with
massive data, being enable to rank observations according to their supposed
degree of abnormality may significantly improve operational processes and allow for a prioritization of actions to be taken, especially in situations where human expertise required to check each observation is time-consuming.
When univariate, observations are usually considered as "abnormal"
when they are  either too high or else too small compared to central
measures such as the mean or the median. In this context, anomaly/novelty analysis generally relies
on the analysis of the tail distribution of the variable of interest.  No natural (pre-) order exists on a $d$-dimensional feature space,  $\mathcal{X}\subset\mathbb{R}^d$ say, as soon as $d>1$. Extension to the multivariate setup is thus far from
obvious and, in practice, the optimal ordering/ranking must be \textit{learned} from training data $X_1,\; \ldots,\; X_n$, in absence of any parametric assumptions on the underlying probability distribution describing the "normal" regime. The most straightforward manner to define a preorder on the feature
space $\mathcal{X}$ is to transport the natural
order on the real half-line through a measurable \textit{scoring
  function} $s:\mathcal{X} \rightarrow \mathbb{R}_+$: the
"smaller" the score $s(X)$, the more "abnormal" the observation $X$ is
viewed. 
Any scoring function defines a preorder on $\mathcal{X}$ and thus a ranking on a set of new observations. An important issue thus concerns the definition of an adequate performance criterion, $\crit(s)$ say, in order to compare possible candidate scoring  function and to pick one eventually: optimal scoring functions $s^*$ being then defined as those optimizing $\crit$. Throughout the present article, it is assumed that the distribution $F$ of the observable r.v. $X$ is absolutely continuous w.r.t. Lebesgue measure $Leb$ on
$\mathcal{X}$, with density $f(x)$.  The criterion should be thus defined  in a way that the collection of level sets of an optimal scoring function $s^*(x)$ coincides with that related to $f$.  In other words, any
nondecreasing transform of the density should be optimal regarding the ranking performance criterion $\crit$. According to the Empirical Risk Minimization (ERM) paradigm, a scoring function will be built in practice by optimizing  an empirical
version $\crit_n(s)$ of the
criterion over an adequate set of scoring functions $\S_0$ of controlled complexity (\textit{e.g.} a major class of finite {\sc VC} dimension). Hence, another desirable property to guarantee the universal consistency of ERM learning strategies is the uniform convergence of
$\crit_n(s)$ to $\crit(s)$  over such collections $\S_0$ under minimal assumptions on the distribution $F(dx)$.
In \cite{CLEM13,CLEM14}, a functional criterion referred to as the Mass-Volume ($\mv$) curve, admissible  with respect to the requirements listed above has been introduced, extending somehow the concept of $\roc$ curve in the unsupervised setup. Relying on the theory of \textit{minimum volume} sets (see \textit{e.g.} \cite{Polonik97,ScottNowak06} and the references therein), it has been proved that the scoring functions minimizing empirical and discretized versions of the $\mv$ curve criterion are accurate when the underlying distribution has compact support and a first algorithm for building nearly optimal scoring functions, based on the estimate of a finite collection of properly chosen minimum volume sets, has been introduced and analyzed. However, by construction, learning rate bounds are rather slow (of the order $n^{-1/4}$ namely) and cannot be established in the unbounded support situation, unless very restrictive assumptions are made on the tail behavior of $F(dx)$. See Figure \ref{algo-problem} and related comments for an insight into the gain resulting from the concept introduced in the present paper in contrast to the $\mv$ curve minimization approach.

 Given these limitations, it is the major goal of this paper to propose an alternative criterion for
anomaly ranking/scoring, called the \textit{Excess-Mass}
  curve ($\EM$ curve in short) here, based on the notion of
{\it density contour clusters}  \cite{Polonik95,hartigan1987estimation,muller1991excess}. Whereas minimum volume sets are solutions of volume minimization problems under mass constraints, the latter are solutions of mass maximization under volume constraints. Exchanging this way objective and constraint, the relevance of this performance measure is thoroughly discussed and accuracy of solutions
which optimize statistical counterparts of this criterion is
investigated. More specifically, rate bounds of the order $n^{-1/2}$ are proved, even in the case of unbounded support. Additionally, in contrast to the analysis carried out in \cite{CLEM13}, the model bias issue is tackled,
insofar as the assumption that the level sets of the underlying
density $f(x)$ belongs to the class of sets used to build the scoring
function is relaxed here. 

 The rest of this paper is organized as follows. Section
\ref{sec:notations} introduces the notion of $\EM$ curve and that of optimal $\EM$
curve. Estimation in the compact support case is covered by section
\ref{sec:estim}, extension to distributions with non compact support and
control of the model bias are tackled in section
\ref{sec:ext}. A simulation study is
performed in section \ref{sec:simul}. All proofs are deferred to the
Appendix section.




\section{Background and related work} \label{sec:background}
As a first go, we first provide a brief overview
of the scoring approach based on  the $\mv$ curve criterion, as a basis
for comparison with that promoted in the present paper.

Here and throughout, the indicator function of any event $\mathcal{E}$ is denoted by $\mathds{1}_{\mathcal{E}}$, the Dirac mass at any point $x$ by $\delta_x$, $A\Delta B$ the symmetric difference between two sets $A$ and $B$ and by $\mathcal{S}$ the set of all scoring functions
$s: \mathcal{X} \rightarrow \mathbb{R}_+ $ integrable w.r.t Lebesgue
measure.
Let $s\in \mathcal{S}$. As defined in \cite{CLEM13,CLEM14}, the
MV-curve of $s$ is the plot of the mapping $\alpha\in (0,1)\mapsto MV_s(\alpha) = \lambda_s \circ \alpha_s^{-1}(\alpha)$,
where $\alpha_s(t)= \mathbb{P}(s(X) \ge t)$, $\lambda_s(t)=Leb(\{x \in \mathcal{X}, s(x) \ge t\})$ and $H^{-1}$ denotes the pseudo-inverse of any cdf $H:\mathbb{R}\rightarrow (0,1)$.
\noindent
This induces a partial ordering on the set of all scoring functions: $s$ is
preferred to $s'$ if $MV_{s}(\alpha) \le MV_{s'}(\alpha)$ for all
$\alpha\in(0,1)$.
One may show that $\mv^*(\alpha)\leq \mv_s(\alpha)$ for all $\alpha\in (0,1)$ and any scoring function $s$, where $MV^*(\alpha)$ is the optimal value of the constrained minimization problem
\begin{equation}\label{eq:MV}\min_{\Gamma~ borelian} ~Leb(\Gamma) \mbox{~subject to~} \mathbb{P}(X \in \Gamma) \ge \alpha.
\end{equation}
Suppose now that $F(dx)$ has a density $f(x)$ satisfying the following assumptions:

\noindent $\mathbf{A_1}$ {\it The density $f$ is bounded, \textit{i.e.} $\vert \vert f(X)\vert\vert_{\infty}<+\infty~.$}
\noindent $\mathbf{A_2}$ {\it The density $f$ has no flat parts: $\forall c\geq 0$, $\mathbb{P}\{f(X)=c\}=0~.$}
 One may then show that the curve $\mv^*$ is actually a $\mv$ curve, that is related to (any increasing transform of) the density $f$ namely: $\mv^*=\mv_f$. In addition, the  minimization problem \eqref{eq:MV} has a unique solution
$\Gamma_\alpha^*$ of mass $\alpha$ exactly, referred to as \textit{minimum volume set} (see \cite{Polonik97}): $\mv^*(\alpha)=Leb(\Gamma^*_\alpha)$ and $F(\Gamma_\alpha^*)=\alpha$. Anomaly scoring can be then viewed as the problem of building a scoring function $s(x)$ based on training data such that $\mv_s$ is (nearly) minimum everywhere, \textit{i.e.} minimizing $\|\mv_{s}-\mv^*\|_{\infty}\overset{def}{=}\sup_{\alpha\in[0,1]}\vert \mv_s(\alpha)-\mv^*(\alpha)\vert$.
Since $F$ is unknown, a minimum volume set estimate $\widehat{\Gamma}^*_{\alpha}$ can be defined as the solution of \eqref{eq:MV} when $F$ is replaced by its empirical version
$F_n=(1/n)\sum_{i=1}^n\delta_{X_i}$, minimization is restricted to a collection $\mathcal{G}$ of borelian subsets of $\mathcal{X}$ supposed not too complex but rich enough to include all density level sets (or reasonable approximants of the latter) and $\alpha$ is replaced by $\alpha-\phi_n$, where the {\it tolerance parameter} $\phi_n$ is a probabilistic upper bound for the supremum $\sup_{\Gamma\in \mathcal{G}}\vert F_n(\Gamma)-F(\Gamma) \vert$. Refer to \cite{ScottNowak06} for further details. The set $\mathcal{G}$ should ideally offer statistical and computational advantages both at the same time. Allowing for fast search on the one hand and being sufficiently complex to capture the geometry of target density level sets on the other.
In \cite{CLEM13}, a method consisting in preliminarily estimating a collection of minimum volume sets related to target masses $0<\alpha_1<\ldots<\alpha_K<1$ forming a subdivision of $(0,1)$ based on training data so as to build a scoring function $s=\sum_k \mathds{1}_{x\in \hat \Gamma_{\alpha_k}^*}$ has been proposed and analyzed. Under adequate assumptions (related to $\mathcal{G}$, the perimeter of the $\Gamma^*_{\alpha_k}$'s and the subdivision step in particular) and for an appropriate choice of $K=K_n$ either under the very restrictive assumption that $F(dx)$ is compactly supported or else by restricting the convergence analysis to $[0,1-\epsilon]$ for $\epsilon>0$, excluding thus the tail behavior of the distribution $F$ from the scope of the analysis, rate bounds of the order $\mathcal{O}_{\mathbb{P}}(n^{-1/4})$ have been established to guarantee the generalization ability of the method.

Figure \ref{algo-problem} illustrates the problems inherent to the use of the $\mv$ curve as a performance criterion for anomaly scoring in a "non asymptotic" context, due to the prior discretization along the mass-axis. In the $2$-d situation described by Fig. \ref{algo-problem} for instance, given the training sample and the partition of the feature space depicted, 
the $\mv$ criterion leads to consider the sequence of empirical minimum volume sets $A_1,\; A_1\cup A_2,\; A_1\cup A_3,\; A_1\cup A_2\cup A_3$ and thus the scoring function $s_1(x)=\mathbb{I}\{x\in A_1  \}+ \mathbb{I}\{x\in A_1\cup A_2  \} + \mathbb{I}\{x\in A_1\cup A_3  \}$, whereas the scoring function $s_2(x)=\mathbb{I}\{x\in A_1  \}+ \mathbb{I}\{x\in A_1\cup A_3  \}$ is clearly more accurate.
\par In this paper, a different functional criterion is proposed, obtained by exchanging objective and constraint functions in \eqref{eq:MV}, and it is shown that optimization of an empirical discretized version of this performance measure yields scoring rules with convergence rates of the order $\mathcal{O}_{\mathbb{P}}(1/\sqrt{n})$. In addition, the results can be extended to the situation where the support of the distribution $F$ is not compact.

\section{The Excess-Mass curve}\label{sec:notations}

The performance criterion we propose in order to evaluate anomaly scoring accuracy relies on 
the notion of \textit{excess mass} and \textit{density contour
clusters}, as introduced in the seminal contribution \cite{Polonik95}. The main idea is to consider a Lagrangian formulation of a constrained minimization problem, obtained by exchanging constraint and objective in \eqref{eq:MV}: for $t>0$,
\begin{equation}
\label{solomeg}
\max_{\Omega~ borelian}  \left\{ \mathbb{P}(X \in \Omega) - t Leb(\Omega) \right\}.
\end{equation}
We denote by $\Omega^*_t$ any solution of this problem. As shall be seen in the subsequent analysis  (see Proposition \ref{propmono}
 below), compared to the $\mv$ curve approach, this formulation offers certain computational and theoretical advantages both at the same time: when letting (a discretized version of) the Lagrangian multiplier $t$ increase from $0$ to infinity, one may easily obtain solutions of empirical counterparts of \eqref{solomeg} forming a \textit{nested} sequence of subsets of the feature space, avoiding thus deteriorating rate bounds by transforming the empirical solutions so as to force monotonicity.
\begin{definition}\label{def:opt}{\sc (Optimal $\EM$ curve)} The optimal Excess-Mass curve related to a given probability distribution $F(dx)$ is defined as the plot of the mapping $$t>0\mapsto \EM^*(t)\overset{def}{=}\max_{\Omega\text{ borelian} } \{ {\mathbb{P}} (X\in \Omega)-tLeb(\Omega) \}.$$ 
\end{definition}
Equipped with the notation above, we have: $EM^*(t)=\mathbb{P}(X \in \Omega_t^*)-t Leb(\Omega_t^*)$ for all $t>0$.
Notice also that $\EM^*(t) = 0$ for any $t>\|f\|_\infty\overset{def}{=}\sup_{x\in \mathcal{X}}\vert f(x)\vert$. 

\begin{figure}[h!]
\centering
\begin{tikzpicture}[scale=0.65]

\newcommand*{\maxx}{10}
\newcommand*{\maxy}{7}
\newcommand*{\maxf}{9}

\draw[->](0,0)--(\maxx,0) node[below]{\scriptsize $t$};
\draw[->](0,0)--(0,\maxy+1) node[left]{\scriptsize $EM^*(t)$}; 

\draw[very thick](\maxf,2pt)--(\maxf,-2pt) node[below]{\scriptsize  $||f||_\infty$} ;
\draw[very thick](2pt,\maxy)--(-2pt,\maxy) node[left]{\scriptsize  1} ;
\draw[very thick](0pt,0)--(0pt,0) node[below]{\small 0} ;

\draw (\maxx/2-0.2,\maxy/2+2.3) node[below]{\scriptsize  Corresponding} ;
\draw (\maxx/2+1,\maxy/2+1.7) node[below]{\scriptsize  distributions $f$ :} ;

\draw[->](2,\maxy-0.5)--(4,\maxy-0.5) node[below]{};
\draw[->](2,\maxy-0.5)--(2,\maxy+1) node[left]{}; 
\draw (2.7,\maxy-0.5) ..controls +(0,1) and +(0,-1).. (2.7,\maxy +0.5);
\draw (2.7,\maxy+0.5) ..controls +(1,0) and +(-1,0).. (3.2,\maxy +0.5);
\draw (3.2,\maxy-0.5) ..controls +(0,1) and +(0,-1).. (3.2,\maxy +0.5);
\draw (3.1,\maxy+1.2) node[below][scale=0.8]{\scriptsize  finite support} ;

\draw[->](5,\maxy-0.5)--(7,\maxy-0.5) node[below]{};
\draw[->](5,\maxy-0.5)--(5,\maxy+1) node[left]{}; 
\draw [blue, dotted, thick] (5.7,\maxy-0.5) ..controls +(0.3,1.2) and +(-0.3,-1.2).. (6,\maxy+0.7);
\draw [blue, dotted, thick] (6,\maxy+0.7) ..controls +(0.3,-1.2) and +(-0.3,1.2).. (6.3,\maxy-0.5);
\draw (6.1,\maxy+1.2) node[below][scale=0.8]{\scriptsize finite support} ;

\draw[->](8,\maxy-0.5)--(10,\maxy-0.5) node[below]{};
\draw[->](8,\maxy-0.5)--(8,\maxy+1) node[left]{}; 
\draw [green, dashed, thick] (8,\maxy-0.5) ..controls +(1,0.1) and +(-0.3,-0.01).. (9,\maxy+0.7);
\draw [green, dashed, thick] (9,\maxy+0.7) ..controls +(0.3,-0.01) and +(-1,0.1).. (10,\maxy-0.5);
\draw (9.3,\maxy+1.2) node[below][scale=0.8]{\scriptsize  infinite support} ;

\draw[->](8,\maxy-3.5)--(10,\maxy-3.5) node[below]{};
\draw[->](8,\maxy-3.5)--(8,\maxy-2) node[left]{}; 
\draw [red,ultra thick] (8,\maxy-3.48) ..controls +(1,0) and +(-0.1,-0.3).. (9,\maxy-2.3);
\draw [red,ultra thick] (9,\maxy-2.3) ..controls +(0.1,-0.3) and +(-1,0).. (10,\maxy-3.48);
\draw (9,\maxy-1.8) node[below][scale=0.8]{\scriptsize  heavy tailed} ;

\newcommand*{\aaa}{(1/sqrt(sqrt(\maxf))+1)}
\newcommand*{\bbb}{(-1/sqrt(sqrt(\maxf)))}

\draw [red,ultra thick,domain=0:\maxf, samples=200] plot (\x,{\maxy * (\aaa/(1+sqrt(sqrt(\x)))+\bbb)} ) ;
\draw [domain=0:\maxf, samples=200] plot (\x,{\maxy * (1- 1/\maxf * \x)} ) ;

\draw [green,dashed, thick] (0,\maxy) ..controls +(0,-6) and +(-1,0).. (\maxf,0); 
\draw [blue, dotted, thick] (0,\maxy) ..controls +(3,-6) and +(-1,0).. (\maxf,0); 

\tikzstyle {tangente} = [very thick, ->];
\draw [tangente,red] (0,\maxy)--++( 0,-1.5);
\draw [tangente,green,dashed] (0,\maxy)--++( 0,-1);
\draw [tangente,blue,dotted] (0,\maxy)--++( 0.5,-1);
\draw [tangente] (0,\maxy)--++( 1,-0.75);


\end{tikzpicture}
\caption{EM curves depending on densities}
\label{EMcurves}
\end{figure}
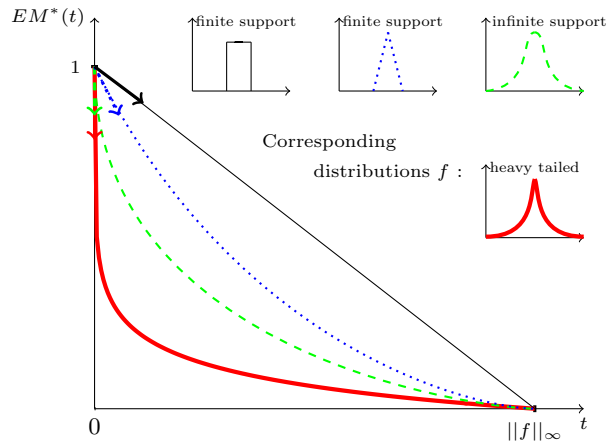

\begin{center}
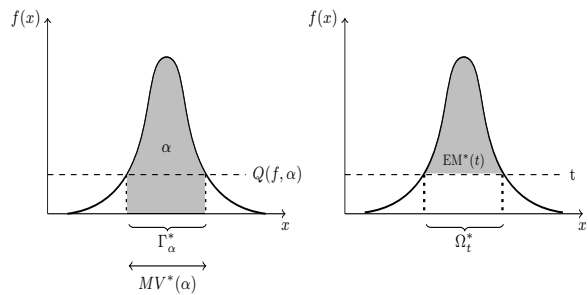
\begin{figure}
\centering
\resizebox{8cm}{4cm}{
\begin{tikzpicture}[scale=1.8]

\draw[->](7.8,0)--(10.2,0) node[below]{\scriptsize $x$};
\draw[->](7.8,0)--(7.8,1.5) node[left]{\scriptsize $f(x)$}; 
\draw [thick] (8,0) ..controls +(1,0.1) and +(-0.3,-0.01).. (9,1.2);
\draw [thick] (9,1.2) ..controls +(0.3,-0.01) and +(-1,0.1).. (10,0);

\draw[dotted,very thick](8.60,0.3)--(8.60,0) node[right]{};
\draw[dotted,very thick](9.39,0.3)--(9.39,0) node[right]{};


\begin{scope} 
\clip (8.61,0)--(9.38,0)--(9.38,1.5)--(8.61,1.5)--(8.61,0) ; 

\path[draw,fill=lightgray] (8,0) ..controls +(1,0.1) and +(-0.3,-0.01).. (9,1.2)--
(9,1.2) ..controls +(0.3,-0.01) and +(-1,0.1).. (10,0)--
(8,0)--(10,0) --cycle;
\end{scope}

\draw[dashed](7.8,0.3)--(9.8,0.3) node[right]{\scriptsize $Q(f,\alpha)$};

\draw[decorate,decoration={brace}]
(9.4,-0.08)--(8.62,-0.08) node[below,pos=0.5] {\scriptsize $\Gamma_\alpha^*$};

\draw[<->]
(9.4,-0.4)--(8.62,-0.4) node[below,pos=0.5] {\scriptsize $MV^*(\alpha)$};

\draw (9,0.5) node[thick]{\scriptsize $\alpha$} ;

\draw[->](10.8,0)--(13.2,0) node[below]{\scriptsize $x$};
\draw[->](10.8,0)--(10.8,1.5) node[left]{\scriptsize $f(x)$}; 
\draw [thick] (11,0.01) ..controls +(1,0.1) and +(-0.3,-0.01).. (12,1.2);
\draw [thick] (12,1.2) ..controls +(0.3,-0.01) and +(-1,0.1).. (13,0.01);
\draw[dashed](10.8,0.3)--(13,0.3) node[right]{\scriptsize t};
\draw[dotted,very thick](11.60,0.3)--(11.60,0) node[right]{};
\draw[dotted,very thick](12.39,0.3)--(12.39,0) node[right]{};


\begin{scope} 
\clip (11,0.31)--(13,0.308)--(13,1.5)--(11,1.5)--(11,0.308) ; 
\path[draw,fill=lightgray] (11,0) ..controls +(1,0.1) and +(-0.3,-0.01).. (12,1.2)--
(12,1.2) ..controls +(0.3,-0.01) and +(-1,0.1).. (13,0)--
(11,0.3)--(13,0.3) --cycle;
\end{scope}

\draw[decorate,decoration={brace}]
(12.4,-0.08)--(11.62,-0.08) node[below,pos=0.5] {\scriptsize  $\Omega_t^*$};

\draw (12,0.4) node[scale=0.85]{\scriptsize $\EM^*(t)$} ;
\draw (11.8,-0.5) node[thick]{ } ;
\end{tikzpicture}
}
\caption{Comparison between $MV^*(\alpha)$ and $EM^*(t)$}
\label{MVcurve}
\end{figure}
\end{center}

\begin{lemma}{\sc (On existence and uniqueness)}
\label{evident}
For any subset $\Omega^*_t$ solution of \eqref{solomeg}, we have
$$\{x, f(x) > t\} \subset \Omega^*_t \subset \{x, f(x) \ge t\} \text{almost-everywhere},$$
 and the sets $\{x, f(x) > t\}$ and $\{x, f(x) \ge t\}$ are both solutions of \eqref{solomeg}.
In addition, under assumption $\mathbf{A_2}$, the solution is unique:
$$\Omega_t^*= \{x, f(x) > t\}= \{x, f(x) \ge t\}.$$
\end{lemma}
 Observe that the curve $\EM^*$ is always well-defined,
since  $ \int_{f \ge t}(f(x)-t)dx = \int_{f > t}(f(x)-t)dx$. We also point out that $\EM^*(t)=\alpha(t)-t\lambda(t)$ for all $t>0$, where we set $\alpha = \alpha_f$ and
$\lambda = \lambda_f$. 
\begin{proposition}{\sc (Derivative and convexity of $\EM^*$)}  Suppose that assumptions $\mathbf{A_1}$ and $\mathbf{A_2}$ are fullfilled. Then, the mapping $\EM^*$ is differentiable and we have for all $t>0$:
\label{derive}
\begin{eqnarray*}
 \EM^{*'}(t)=-\lambda(t). 
\end{eqnarray*}
In addition, the mapping $t>0 \mapsto \lambda(t)$ being decreasing, the curve $EM^*$ is convex.
\end{proposition}

We now introduce the concept of Excess-Mass curve of a scoring function $s\in \S$.
\begin{definition} {\sc ($\EM$ curves)}
The  $\EM$ curve of $s\in\mathcal{S}$  w.r.t. the probability
distribution $F(dx)$ of a random variable $X$ is the plot of the mapping
\begin{equation}
\label{EM}
\EM_s : t \in [0, \infty[ \mapsto \sup_{A \in \{(\Omega_{s,l})_{l>0}\}} {\mathbb{P}}(X \in A) - t Leb(A),
\end{equation}
where $\Omega_{s,t}=\{ x \in \mathcal{X}, s(x) \ge t \}$ for all $t>0$.
One may also write: $\forall t>0$, $\EM_s(t)= \sup_{u>0}~ \alpha_s(u) -
t \lambda_s(u) $. Finally, under assumption $\mathbf{A_1}$, we have $\EM_s(t)=0$ for every $t> \|f\|_\infty$. 
\end{definition}

%
%
%
%
%
%

Regarding anomaly scoring, the concept of $\EM$ curve naturally induces a partial order on the set of all scoring functions: $\forall (s_1,s_2)\in \S^2$, $s_1$ is said to be more accurate than $s_2$ when $\forall t > 0, \EM_{s_1}(t) \geq \EM_{s_2}(t)$. Observe also that the optimal $\EM$ curve introduced in Definition \ref{def:opt} is itself the $\EM$ curve of a scoring function, the $\EM$ curve of any strictly increasing transform of the density $f$ namely: $\EM^*=\EM_f$. Hence, in the unsupervised framework, optimal scoring functions are those maximizing the $\EM$ curve everywhere. In addition, maximizing $\EM_s$ can be viewed as recovering a collection of subsets $(\Omega^*_t)_{t>0}$ with maximum mass when penalized by their volume in a linear fashion. An optimal scoring function is then any $s\in \S$  with the $\Omega^*_t$'s as level sets, for instance any scoring function of the form
\noindent
\begin{align}\label{score_cont}
s(x)=\int_{t=0}^{+\infty} \mathds{1}_{x\in \Omega^*_t}a(t)dt,
\end{align} 
with $a(t)>0$ (observe that $s(x)=f(x)$ for $a \equiv 1$).

\begin{proposition}
\label{propestim} ({\sc Nature of anomaly scoring}) Let $s \in \mathcal{S}$. The following properties hold true.
\begin{enumerate}
\item[(i)] The mapping $\EM_s$ is non increasing on $(0,+\infty)$, takes its values in $[0,1]$ and satisfies,
$\EM_s(t) \le \EM^*(t)$ for all $t\geq 0$. 
\item[(ii)]  For $t \ge 0$, we have: $0 \le \EM^*(t)-\EM_s(t) \le \|f\|_\infty \inf_{u>0} Leb (\{ s >u\}\Delta\{f>t\}).$
\item[(iii)] Let $\epsilon >0$. Suppose that the quantity $\sup_{u>\epsilon}
  \int_{f^{-1}(\{u\})} 1/\|\nabla f(x)\|\;  d\mu(x) $ is bounded,
  where $\mu$ denotes the $(d-1)$-dimensional Hausdorff measure. Set $\epsilon_1 := \inf_{T} \|f-T\circ s\|_\infty$, where the infimum is taken over the set $\mathcal{T}$ of all borelian increasing transforms $T : \mathbb{R}_+ \rightarrow \mathbb{R}_+$. Then 
\begin{align*}
&\sup_{t\in[\epsilon + \epsilon_1,\|f\|_\infty]}|\EM^*(t)-\EM_s(t)| \\ 
&~~~~~~~~~~~~~~~~~~~~~~~~~~~~~~\le  C_1 \inf_{T  \in \mathcal{T}} \|f-T\circ s\|_\infty
\end{align*}
where $C_1=C(\epsilon_1,f)$ is a constant independent from $s(x)$.
\end{enumerate}
\end{proposition}

Assertion $(ii)$ provides a control of  the pointwise difference between the
optimal $\EM$ curve and $\EM_s$  in terms of the error made when recovering a specific minimum volume set $\Omega_t^*$ by a level set of $s(x)$.
Assertion $(iii)$ reveals that, if a certain increasing transform of a given scoring function $s(x)$ approximates well the density $f(x)$, then $s(x)$ is an accurate scoring function w.r.t. the $\EM$ criterion. 
As the distribution $F(dx)$ is generally unknown, $\EM$ curves must be estimated. Let $s\in \S$ and $X_1,\; \ldots,\; X_n$ be an i.i.d. sample with common distribution $F(dx)$ and set $\widehat{\alpha}_s(t)=(1/n)\sum_{i=1}^n\mathds{1}_{s(X_i)\geq t}$. The empirical $\EM$ curve of $s$ is then defined as $$\widehat{\EM}_s(t)=\sup_{u>0}\{ \widehat{\alpha}_s(u)-t\lambda_s(u)\}~.$$ In practice, it may be difficult to estimate the volume $\lambda_s(u)$ and Monte-Carlo approximation can naturally be used for this purpose.

\section{A general approach to learn a scoring function}\label{sec:estim}

The concept of $\EM$-curve provides a simple way to compare scoring functions but optimizing such a functional criterion is far from straightforward. As in \cite{CLEM13}, we propose to discretize the continuum of optimization problems and to construct a nearly optimal scoring function with level sets built by solving a finite collection of empirical versions of problem \eqref{solomeg}
over a subclass $\mathcal{G}$ of borelian subsets. In order to analyze the accuracy of this approach, we introduce the following additional assumptions.

\noindent $\mathbf{A_3}$ {\it All minimum volume sets belong to $\mathcal{G}$: $$\forall t >0,~ \Omega_t^* \in \mathcal{G}~.$$} 

\noindent $\mathbf{A_4} $ {\it The Rademacher average
 $$\mathcal{R}_n=\mathbb{E} \left[ \sup_{\Omega \in \mathcal{G}}
    \frac{1}{n} \left| \sum_{i=1}^n \epsilon_i \mathds{1}_{X_i \in
        \Omega} \right| \right]$$ is of order $\mathcal{O}_{\mathbb{P}}(n^{-1/2})$, where $(\epsilon_i)_{i \ge 1}$ is a Rademacher chaos independent of the $X_i$'s.}
        
Assumption $\mathbf{A_4}$ is very general and is fulfilled in particular when $\mathcal{G}$ is of finite VC dimension, see \cite{Kolt06}, whereas the zero bias assumption $\mathbf{A_3}$ is in contrast very restrictive. It will be relaxed in section~\ref{sec:ext}. 

Let $\delta\in (0,1)$ and consider the complexity penalty $\Phi_n(\delta)=2 \mathcal{R}_n + \sqrt{\frac{log(1/\delta)}{2n}}$. We have for all $n \ge 1$:
\begin{equation}
\label{penality}
\mathbb{P}\left( \left\{ \sup_{G\in \mathcal{G}}\left( |P(G)- P_n(G)|-\Phi_n(\delta) \right) >0 \right\}\right) \leq \delta,
\end{equation}
see \cite{Kolt06} for instance. Denote by $F_n=(1/n)\sum_{i=1}^n \delta_{X_i}$ the empirical measure based on the training sample $X_1,\; \ldots,\; X_n$. 
For $t \ge 0$, define also the signed measures:
\begin{align*}
&H_t(\point) =F(\point)-t Leb(\point)\\ 
\text{and ~~~~} &H_{n,t}(\point)=F_n(\point)-t Leb(\point).
\end{align*}
Equipped with these notations, for any $s \in \mathcal{S}$, we point out that one may write $\EM^*(t)=\sup_{u \ge 0} H_t(\{x \in {\mathcal{X}}, f(x) \ge u\})$ and $\EM_s(t)=\sup_{u \ge 0} H_{t}(\{x \in {\mathcal{X}}, s(x) \ge u\})$. Let $K>0$ and  $0<t_K<t_{K-1}<\ldots<t_1$. For $k$ in $\{1,\; \ldots,\; K\}$, let $\hat \Omega_{t_k}$ be an \textit{empirical $t_k$-cluster}, that is to say a borelian subset of $\mathcal{X}$ such that
$$\hat \Omega_{t_k} \in arg\max_{\Omega \in \mathcal{G}} H_{n,t_k}(\Omega).$$
The empirical excess mass at level $t_k$ is then $H_{n,t_k}(\hat \Omega_{t_k})$. The following result reveals the benefit of viewing density level sets as solutions of \eqref{solomeg} rather than solutions of \eqref{eq:MV} (corresponding to a different parametrization of the thresholds).

\begin{proposition} 
\label{propmono}{\sc (Monotonicity)}
For any $k$ in $\{1,~\ldots,~K\}$, the subsets $\cup_{i \le k}\hat \Omega_{t_i}$ and $\cap_{i \ge k} \hat \Omega_{t_i}$ are still empirical $t_k$-clusters, just like $\hat \Omega_{t_k}$: 
\begin{align*} 
H_{n,t_k}(\cup_{i \le k}\hat \Omega_{t_i})=H_{n,t_k}(\cap_{i \ge k}\hat \Omega_{t_i})=H_{n,t_k}(\hat \Omega_{t_k}).
\end{align*}
\end{proposition}

The result above shows that monotonous (regarding the inclusion) collections of empirical clusters can always be built. Coming back to the example depicted by Fig. \ref{algo-problem}, as $t$ decreases, the $\hat \Omega_{t}$'s are successively equal to $A_1$,~ $A_1 \cup A_3$,~ and $A_1 \cup A_3 \cup A_2$, and are thus monotone as expected. This way, one fully avoids the problem inherent to the prior specification of a subdivision of the mass-axis in the $\mv$-curve minimization approach (see the discussion in section \ref{sec:background}).

Consider an increasing sequence of empirical $t_k$ clusters $(\hat \Omega_{t_k})_{1\leq k\leq K}$ and a scoring function $s \in S$ of the form
\noindent
\begin{align}
\label{score}
s_K(x):= \sum_{k=1}^K a_k \mathds{1}_{x \in \hat{\Omega}_{t_k} }~,
\end{align}
where $a_k>0$ for every $k\in\{1,\; \ldots,\; K\}$. Notice that the scoring function \eqref{score} can be seen as a Riemann sum approximation of \eqref{score_cont} when $a_k=a(t_k)-a(t_{k+1})$. For simplicity solely, we take $a_k=t_{k}-t_{k+1}$ so that the $\hat \Omega_{t_k}$'s  are $t_k$-level sets of $s_K$, \textit{i.e} $\hat \Omega_{t_k}=\{s \ge t_k\}$ and $\{s \ge t\}= \hat \Omega_{t_k}$ if $t \in ]t_{k+1},t_{k}]$. Observe that the results established in this paper remain true for other choices. In the asymptotic framework considered in the subsequent analysis, it is stipulated that $K=K_n \rightarrow \infty$ as $n\rightarrow +\infty$. We assume in addition that $\sum_{k=1}^{\infty}a_k < \infty$.
\begin{remark}
\label{orderscore}{\sc (Nested sequences)}
For $L \le K$, we have $\{\Omega_{s_L,l},l \ge 0\}=(\hat \Omega_{t_k})_{0 \le k \le L } \subset (\hat \Omega_{t_k})_{0 \le k \le K }=\{\Omega_{s_K,l},l \ge 0\}$, so that by definition, $EM_{s_L} \le EM_{s_K}$.   
\end{remark}
\begin{remark}{\sc (Related work)}
  We point out that a very similar result is proved in \cite{polonik1998} (see Lemma 2.2 therein) concerning the
  Lebesgue measure of the symmetric differences of density clusters.
\end{remark}
\begin{remark}{\sc (Alternative construction)}
\label{mono}
It is noteworthy that, in practice, one may solve the optimization problems
$\tilde
\Omega_{t_k} \in \arg\max_{\Omega \in \mathcal{G}} H_{n,t_k}(\Omega)$
and  next form  $\hat \Omega_{t_k}= \cup_{i \le k} \tilde \Omega_{t_i}$. 
\end{remark}

The following theorem provides rate bounds describing the performance of the scoring function $s_K$ thus built with respect to the $\EM$ curve criterion
 in the case where the density $f$ has compact support.
\begin{theorem}{\sc (Compact support case)}
\label{compact_support_case}
Assume that conditions $\mathbf{A_1}$, $\mathbf{A_2}$, $\mathbf{A_3}$ and $\mathbf{A_4}$ hold true, and that $f$ has a compact support. Let $\delta \in ]0,1[$, let
 $(t_k)_{k\in\{1,\;\ldots,\; K\}}$ be such that  $\sup_{1\leq k< K}(t_{k}-t_{k+1}) = \mathcal{O}(1/\sqrt{n})$.
Then, there exists a constant $A$ independent from the $t_k$'s, $n$ and
$\delta$ such that, with probability at least  $1-\delta$, we have:
\begin{align*}
&\sup_{t \in ]0,t_1]} |\EM^*(t)-\EM_{s_K}(t)| \\ &~~~~~~~~~~~~~\le~\left(A+\sqrt{2 \log(1/\delta)}+Leb(supp f)\right)\frac{1}{\sqrt{n}}.
\end{align*}
\end{theorem}

\begin{remark}
\label{supf}{\sc (Localization)}
The problem tackled in this paper is that of scoring anomalies, which correspond to observations lying 
 outside of "large" excess mass sets, namely density clusters with parameter $t$
close to zero. It is thus essential to establish rate bounds 
for the quantity $\sup_{t \in ]0,C[} |\EM^*(t)-\EM_{s_K}(t)|$, where $C>0$ depends
on the proportion of the "least normal" data we want to score/rank. 
\end{remark}

\section{Extensions - Further results}\label{sec:ext}
This section is devoted to extend the results of the previous one. We first relax the compact support assumption and next the one stipulating that all density level sets belong to the class $\mathcal{G}$, namely $\mathbf{A_3}$.

\subsection{Distributions with non compact support}\label{sec:infiniteSupport}
\label{extension_non_compact}
It is the purpose of this section to show that the algorithm detailed below produces a scoring function $s$ such that  $EM_s$ is uniformly close to $EM^*$ (Theorem \ref{thmprinc}). 
 See Figure \ref{algo-problem} as an illustration and a comparaison with the $MV$ formulation as used as a way to recover empirical minimum volume set $\hat \Gamma_\alpha$ .

\begin{algo} \label{algo1} Suppose that assumptions $\mathbf{A_1}$, $\mathbf{A_2}$, $\mathbf{A_3}$, $\mathbf{A_4}$ hold true. Let $t_1$ such that $\max_{\Omega \in \mathcal{G}}H_{n,t_1}(\Omega)
\ge  0$. 
Fix  $N>0$. 
For $k=1,\; \ldots,\; N$, 
\begin{enumerate} 
\item   Find  $\tilde{\Omega}_{t_k} \in \arg\max_{\Omega \in \mathcal{G}}
H_{n,t_k}(\Omega)$ , 
\item  Define $\hat \Omega_{t_k}= \cup_{i \le k} \tilde \Omega_{t_i}$
\item  Set  $ t_{k+1} =\frac{t_1}{(1+\frac{1}{\sqrt n})^{k}}   $ for $k\le N-1$. 
\end{enumerate} 

In order to reduce the complexity, we may replace steps $1$ and $2$
with $\hat {\Omega}_{t_k} \in \arg\max_{\Omega \supset \hat \Omega_{t_{k-1}}} H_{n,t_k}(\Omega)$.
\noindent
The resulting piecewise constant scoring function is
\begin{align}
\label{definition_sN}
s_N(x)= \sum_{k=1}^{N}(t_{k}-t_{k+1}) \mathds{1}_{x \in \hat{\Omega}_{t_k}}~.
\end{align}
\end{algo}

\begin{center}
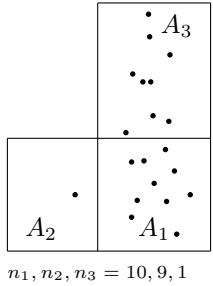
\begin{figure}[h!]
\centering
\begin{tikzpicture}[scale=1.5]

\draw[-](0.2,0)--(2,0) ;
\draw[-](0.2,0)--(0.2,1) ; 
\draw[-](1,0)--(1,2.2) ;
\draw[-](2,0)--(2,2.2) ;
\draw[-](0.2,1)--(2,1) ;
\draw[-](1,2.2)--(2,2.2) ;

\draw (0.8,0.5) node{\Huge .} ;
\draw (1.35,0.45) node{\Huge .} ;
\draw (1.3,0.79) node{\Huge .} ;
\draw (1.25,1.05) node{\Huge .} ;
\draw (1.31,1.57) node{\Huge .} ;
\draw (1.45,2.1) node{\Huge .} ;
\draw (1.47,1.5) node{\Huge .} ;
\draw (1.41,0.8) node{\Huge .} ;
\draw (1.49,1.2) node{\Huge .} ;
\draw (1.46,1.9) node{\Huge .} ;
\draw (1.4,1.5) node{\Huge .} ;

\draw (1.61,0.44) node{\Huge .} ;
\draw (1.68,0.71) node{\Huge .} ;
\draw (1.63,1.15) node{\Huge .} ;
\draw (1.64,1.74) node{\Huge .} ;
\draw (1.5,0.6) node{\Huge .} ;
\draw (1.7,0.15) node{\Huge .} ;
\draw (1.3,0.3) node{\Huge .} ;
\draw (1.82,0.5) node{\Huge .} ;
\draw (1.6,0.9) node{\Huge .} ;

\draw (1.5,0.2) node{$A_1$} ;
\draw (0.5,0.2) node{$A_2$} ;
\draw (1.7,2.) node{$A_3$} ;

\draw (1,-0.2) node{\scriptsize $n_1,n_2,n_3=10,9,1$} ;
\end{tikzpicture}
\caption{ \footnotesize Sample of $n=20$ points in a $2$-d space, partitioned into three rectangles.  As $\alpha$ increases, the minimum volume sets $\hat \Gamma_{\alpha}$ are successively equal to $A_1$,~ $A_1 \cup A_2$,~ $A_1 \cup A_3$, and $A_1 \cup A_3 \cup A_2$, whereas, in the $EM$-approach, as $t$ decreases, the $\hat \Omega_{t}$'s are successively equal to $A_1$,~ $A_1 \cup A_3$,~ and $A_1 \cup A_3 \cup A_2$.
}

\label{algo-problem}
\end{figure}
\end{center}

The main argument to extend the above results to the case where $supp f$ is not bounded is given in Lemma \ref{theofini} in the "Technical Details" section. The meshgrid $(t_k)$ must be chosen adaptively, in a data-driven fashion.  Let $h:~\mathbb{R}_+^* \rightarrow \mathbb{R}_+$ be a decreasing function such that $ \lim_{t \rightarrow 0}h(t)=+\infty$. Just like the previous approach, the grid is described by a decreasing sequence $(t_k)$. Let $t_1 \ge 0$, $N>0$ and define recursively $t_1>t_{2}>\ldots>t_{N}>t_{N+1}=0$, as well as $\hat \Omega_{t_1},\; \ldots,\;\hat \Omega_{t_N}$, through
\noindent
\begin{align}
\label{tk-omegak}t_{k+1}&~=~t_k -(\sqrt n)^{-1}\frac{1}{h(t_{k+1})}  \\ 
\label{tk1} \hat{\Omega}_{t_k}&~=~\arg\max_{\Omega \in \mathcal{G}} H_{n,t_k}(\Omega),
\end{align}

with the property that $\hat \Omega_{t_{k+1}} \supset \hat \Omega_{t_k}$. As pointed out in Remark \ref{mono}, it suffices to take $\hat \Omega_{t_{k+1}}=\tilde \Omega_{t_{k+1}} \cup \hat \Omega_{t_{k}}$, where $\tilde{\Omega}_{t_{k+1}}=\arg\max_{\Omega \in \mathcal{G}} H_{n,t_k}(\Omega)$.
This yields the scoring function $s_N$ defined by (\ref{definition_sN}) such that by virtue of Lemma \ref{theofini} (see the Technical Deails), with probability at least $1-\delta$,
\begin{align*}
&\sup_{t \in ]t_N,t_1]} |EM^*(t)-EM_{s_N}(t)| \\ &~~~~~~~~~\le~\left(A+\sqrt{ 2 \log(1/\delta)}~+~\sup_{1\leq k\leq N}\frac{\lambda(t_k)}{h(t_k)} \right)\frac{1}{\sqrt{n}}~. 
\end{align*}
\noindent
Therefore, if we take $h$ such that $\lambda(t) = \mathcal{O}(h(t))$
as $t \rightarrow 0$, we can assume that $\lambda(t)/h(t) \le B$ for t in $]0,t_1]$ since $\lambda$ is decreasing, and we
obtain:
\begin{align}
\label{fondineq}
& \sup_{t \in ]t_N,t_1]} |EM^*(t)-EM_{s_N}(t)| \nonumber \\&~~~~~~~~~~~~~~~\le~\left(A+\sqrt{2 \log(1/\delta)} \right)\frac{1}{\sqrt{n}}~.
\end{align}
\noindent
On the other hand from $t Leb(\{f>t\}) \le \int_{f>t} f \le 1$, we have $\lambda(t) \le 1/t$. Thus $h$ can be chosen as $h(t):=1/t$ for $t \in ]0,t_1]$. In this case, (\ref{tk1}) yields, for $k \ge 2$,
\begin{align}
\label{tk}
t_{k}=\frac{t_1}{(1+\frac{1}{\sqrt n})^{k-1}}~.
\end{align}

\begin{theorem}({\sc Unbounded support case})
\label{thmprinc}
Suppose that assumptions $\mathbf{A_1}$, $\mathbf{A_2}$, $\mathbf{A_3}$, $\mathbf{A_4}$ hold true, let $t_1>0$ and for $k \ge 2$, consider $t_k$ as defined by \eqref{tk}, $\Omega_{t_k}$ by (\ref{tk-omegak}), and $s_N$ (\ref{definition_sN}). Then there is a constant $A$ independent from $N$, $n$ and $\delta$ such that, with probability larger than $1-\delta$, we have: 
\begin{align*}
&\sup_{t \in ]0,t_1]}|EM^*(t)-EM_{s_N}(t)| \\ &~~~~~~~~~~~~~~~~~~~~~\le~ \left[A+\sqrt{2\log(1/\delta)}\right]\frac{1}{\sqrt n} + o_N(1),
\end{align*}
where $o_N(1)=1-EM^*(t_N)$. In addition, $s_N(x)$ converges to $s_\infty(x):=\sum_{k=1}^{\infty} (t_{k+1}-t_k)\mathds{1}_{\hat \Omega_{t_{k+1}}}$ as $N \rightarrow \infty$ and $s_\infty$ is such that, for all $\delta \in (0,1)$, we have with probability at least $1-\delta$:
\begin{align*}
\sup_{t \in ]0,t_1]}|EM^*(t)-EM_{s_\infty}(t)| \le \left[A+\sqrt{2\log(1/\delta)}\right]\frac{1}{\sqrt n}
\end{align*}
\end{theorem}

\subsection{Bias analysis}
\label{biais}
In this subsection, we relax assumption $\mathbf{A_3}$. For any collection $\mathcal{C}$ of subsets of $\mathbb{R}^d$, $\sigma(\mathcal{C})$ denotes here the $\sigma$-algebra generated by $\mathcal{C}$.
Consider the hypothesis below.

\noindent
$\mathbf{\tilde A_3}$  {\it There exists a countable subcollection of $\mathcal{G}$, $F=\{F_i\}_{i \ge 1}$ say, forming a partition of $\mathcal{X}$ and such that $\sigma (F) \subset \mathcal{G}$.} 

 Denote by $f_{F}$ the best approximation (for the $L_1$-norm) of $f$ by piecewise functions on $F$, $$f_{F}(x) := \sum_{i \ge 1} \mathds{1}_{x \in F_i} \frac{1}{Leb(F_i)} \int_{F_i}f(y)dy~.$$
Then, variants of Theorems \ref{compact_support_case} and \ref{thmprinc} can be established without assumption $\mathbf{A_3}$, as soon as $\mathbf{\tilde A_3}$ holds true, at the price of the additional term $\|f-f_{F}\|_{L^1}$ in the bound, related to the inherent bias. For illustration purpose, the following result generalizes one of the inequalities stated in Theorem \ref{thmprinc}:

\begin{theorem}({\sc Biased empirical clusters})
\label{thmprincbiais}
Suppose that assumptions $\mathbf{A_1}$, $\mathbf{A_2}$,
$\mathbf{\tilde A_3}$, $\mathbf{A_4}$ hold true, let $t_1>0$ and for $k \ge 2$ consider $t_k$ defined by (\ref{tk}), $\Omega_{t_k}$ by (\ref{tk-omegak}), and $s_N$ by (\ref{definition_sN}). Then there is a constant $A$ independent from $N$, $n$, $\delta$ such that, with probability larger than $1-\delta$, we have: 
\begin{align*}
&\sup_{t \in ]0,t_1]}|EM^*(t)-EM_{s_N}(t)| \\ &~~~~\le~ \left[A+\sqrt{2\log(1/\delta)}\right]\frac{1}{\sqrt n} + \|f-f_{F}\|_{L^1}  + o_N(1), 
\end{align*}
where $o_N(1)=1-EM^*(t_N)$. 
\end{theorem}

\begin{remark}{\sc (Hypercubes)}
In practice, one defines a sequence of models $F_l\subset\mathcal{G}_l$ indexed by a tuning parameter $l$ controlling (the inverse of) model complexity, such that $\|f-f_{F_l}\|_{L^1} \rightarrow 0$ as $l \rightarrow 0$. 
For instance, the class $F_l$ could be formed by disjoint hypercubes of side length $l$.
\end{remark}

\section{Simulation examples}
\label{sec:simul}
\begin{figure}[htbp]
\begin{minipage}[c]{.45\linewidth}
\begin{center}
\includegraphics[width=\linewidth,height=6.6cm]{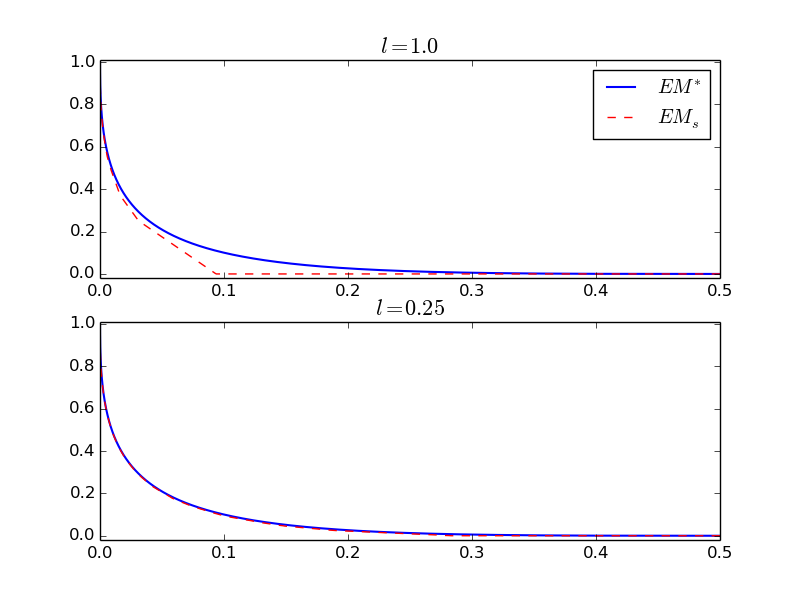}
\caption{Optimal and realized EM curves} 
\label{EMMS}
\end{center}
\end{minipage}
\hfill
\begin{minipage}[c]{.45\linewidth}
\begin{center}
\includegraphics[width=\linewidth,height=6.6cm]{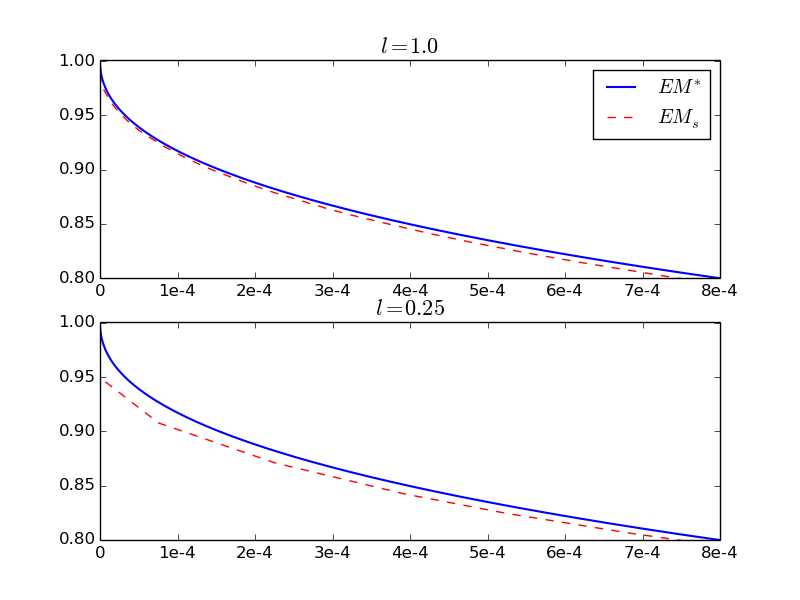}
\caption{Zoom near 0 ~~~~~~~~~~~~~~~~~~~~~~~~~~~~~~~~~~~~~~~~~~~~~~~~~} 
\label{EMMSzoom}
\end{center}
\end{minipage}
\end{figure}

\begin{figure}[!ht]
\includegraphics[width=\linewidth,height=5.8cm]{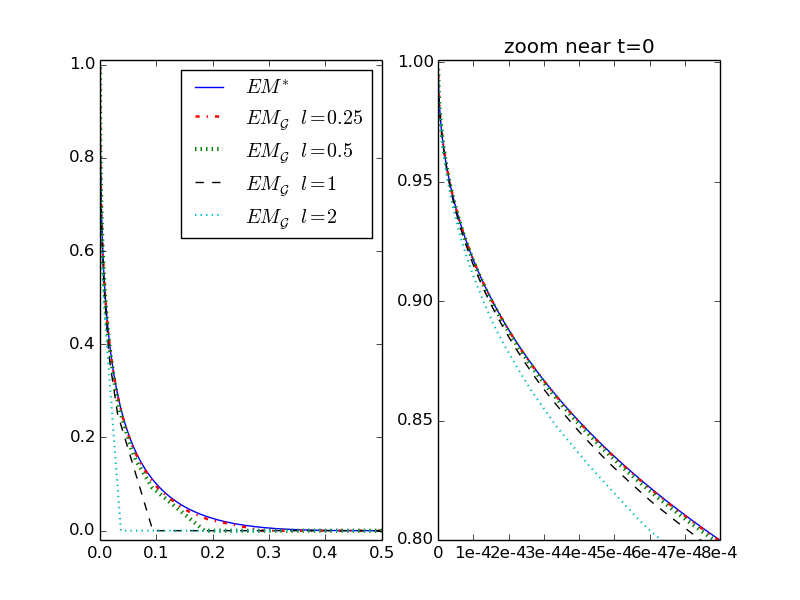}
\caption{$EM_\mathcal{G}$ for different $l$} 
\label{EMGEM}
\end{figure}

Algorithm \ref{algo1} is here implemented from simulated $2$-$d$ \textit{heavy-tailed} data  with common density
$f(x,y)=1/2 \times 1/(1+|x|)^3\times 1/(1+|y|)^2$. The training set is of size $n=10^5$, whereas the test set counts $10^6$ points.
For $l>0$, we set $\mathcal{G}_l=\sigma(F)$ where $F_l=\{F_i^l\}_{i\in \mathbb{Z}^2}$ and $F_i^l = [l i_1,l i_1+1]\times[l i_2,l i_2+1]$ for all $i=(i_1,i_2) \in \mathbb{Z}^2$. The bias of the model is thus bounded by $\|f-f_{F}\|_{\infty}$, vanishing as $l \rightarrow 0$ (observe that the bias is at most of order $l$ as soon as $f$ is Lipschitz for instance). The scoring function $s$ is built using the points located in $[-L,L]^2$ and setting $s=0$ outside of $[-L,L]^2$. Practically, one takes $L$ as the maximum norm value of the points in the training set, or such that an empirical estimate of $\mathbb{P}(X \in [-L,L]^2)$ is very close to $1$ (here one obtains $0.998$ for $L=500$). 
The implementation of our algorithm involves the use of a sparse matrix to store the data in the partition of hypercubes, such that the complexity of the procedure for building the scoring function $s$ and that of the computation of its empirical $\EM$-curve is very small compared to that needed to compute $f_{F_l}$ and $EM_{f_{F_l}}$, which are given here for the sole purpose of quantifying the model bias.\\
Fig. \ref{EMMS} illustrates as expected the deterioration of $EM_s$ for large $l$, except for $t$ close to zero: this corresponds to the model bias. However, Fig. \ref{EMMSzoom} reveals an "overfitting" phenomenon for values of $t$ close to zero, when $l$ is fairly small. 
This is mainly due to the fact that subsets involved in the scoring function are then tiny in regions where there are very few observations (in the tail of the distribution). On the other hand, for the largest values of $t$, the smallest values of $l$ give the best results: the smaller the parameter $l$, the weaker the model bias and no overfitting is experienced because of the high local density of the observations.
Recalling the notation $EM_{\mathcal{G}}^*(t)=\max_{\Omega \in \mathcal{G}}H_t(\Omega) \le EM^*(t)=\max_{\Omega\; meas.}H_t(\Omega)$ so that the bias of our model is $EM^*-EM^*_{\mathcal{G}}$, Fig. \ref{EMGEM} illustrates the variations of the bias with the wealth of our model characterized by $l$ the width of the partition by hypercubes. Notice that partitions with small $l$ are not so good approximation for large $t$, but are performing as well as the other in the extreme values, namely when $t$ is close to $0$. On the top of that, those partitions have the merit not to overfit the extreme datas, which typically are isolated.
 
This empirical analysis demonstrates that introducing a notion of adaptivity for the partition $F$, with progressively growing bin-width as $t$ decays to zero and as the hypercubes are being selected in the construction of $s$ (which crucially depends on local properties of the empirical distribution), drastically improves the accuracy of the resulting scoring function in the $\EM$ curve sense. 
\noindent

\section{Conclusion}
Prolongating the contribution of \cite{CLEM13}, this article provides an alternative view (respectively, an other parameterization) of the anomaly scoring problem, leading to another adaptive method to build scoring functions, which offers theoretical and computational advantages both at the same time. This novel formulation yields a procedure producing a nested sequence of empirical density level sets, and exhibits a good performance, even in the non compact support case. In addition, the model bias has been incorporated in the rate bound analysis. 

\section*{Technical Details}
\textbf{Proof of Theorem \ref{compact_support_case} (Sketch of)} The proof results from the following lemma, which does not use the compact support assumption on $f$ and is the starting point of the extension to the non compact support case (section \ref{extension_non_compact}).
\begin{lemma} 
\label{theofini} Suppose that assumptions $\mathbf{A_1}$, $\mathbf{A_2}$, $\mathbf{A_3}$ and
  $\mathbf{A_4}$ are fulfilled. Then, for $1 \le k \le K-1$, there exists a constant $A$ independent from $n$ and  $\delta$, such that, with probability at least $1-\delta$, for $t$ in $]t_{k+1},t_{k}]$, 
\begin{align*}
|\EM^*(t)-\EM_{s_K}(t)| ~\le~&\left(A+\sqrt{2log(1/\delta)}\right)\frac{1}{\sqrt n}\\ &~~~~~+~ \lambda(t_{k+1})(t_{k}-t_{k+1}).
\end{align*}
\end{lemma}
The detailed proof of this lemma is in the supplementary material, and is a combination on the two following results, the second one being a straightforward consequence of the derivative property of $EM^*$ (Proposition~\ref{derive}):
\begin{itemize}
\item With probability at least $1-\delta$, for $k \in \{1,...,K\}$, $$0 \le EM^*(t_{k})-EM_{s_K}(t_k) \le 2 \Phi_n(\delta)~.$$ 
\item Let $k$ in $\{1,...,K-1\}$. Then for every $t$ in $]t_{k+1},t_{k}]$,
\begin{align*}
0 \le EM^*(t)-EM^*(t_{k}) \le \lambda(t_{k+1}) (t_{k}-t_{k+1})~.
\end{align*}
\end{itemize}

\textbf{Proof of Theorem \ref{thmprinc} (Sketch of)}
The first assertion is a consequence of \eqref{fondineq} combined with the fact that
\begin{align*}
&\sup_{t \in ]0,t_N]} |EM^*(t)-EM_{s_N}(t)| ~\leq~ 1-EM_{s_N}(t_N) \\ &~~~~~~~~~~~~~~~~~~~~~~~~~~~~~~~~~~\leq~ 1-EM^*(t_N)+2\Phi_n(\delta)
\end{align*} 
holds true with probability at least $1-\delta$.
\noindent For the second part, it suffices to observe that $s_N(x)$ (absolutely) converges to $s_{\infty}$ and that, as pointed out in Remark \ref{orderscore}, $EM_{s_N} \le EM_{s_\infty}$. 

\textbf{Proof of Theorem \ref{thmprincbiais} (Sketch of)}
The result directly follows from the following lemma, which establishes an upper bound for the bias, with the notations $\EM_{\mathcal{C}}^*(t):=\max_{\Omega \in \mathcal{C}}H_t(\Omega) \le \EM^*(t)=\max_{\Omega\; meas.}H_t(\Omega)$ for any class of measurable sets $\mathcal{C}$, and $\mathcal{F}:= \sigma(F)$ so that by assumption $\mathbf{A_3}$, $\mathcal{F} \subset \mathcal{G}$. Details are omitted due to space limits.\\

\begin{lemma} 
\label{propbiais}
Under assumption $\mathbf{\tilde A_3}$,  we have for every $t$ in $[0,\|f\|_\infty]$, $$0\le \EM^*(t)-\EM^*_{\mathcal{F}}(t) \le \|f-f_{F}\|_{L^1}~. $$ The model bias $\EM^*-\EM^*_{\mathcal{G}}$ is then uniformly bounded by $\|f-f_{F}\|_{L^1}$.
\end{lemma}

To prove this lemma (see the supplementary material for details), one shows that:
\begin{align*}
\EM^*(t)-\EM^*_{\mathcal{F}}(t) &\le \int_{f>t}(f-f_{F}) \\ &~~+ \int_{\{f>t\}\setminus \{f_{F}>t\}}(f_{F}-t) \\ &~~-\int_{\{f_{F}>t\} \setminus \{f>t\}}(f_{F}-t)~,
\end{align*}
where we use the fact that for all $t>0$, $\{ f_{F} > t \} \in \mathcal{F}$ and $\forall F~ \in~ \mathcal{F},~ \int_Gf=\int_Gf_{F}$.
It suffices then to observe that the second and the third term in the bound are non-positive.
\newpage

\subsubsection*{References}
\renewcommand\refname{\vskip -1cm} 
\bibliographystyle{plain}
\bibliography{mvset}

\begin{thebibliography}{10}

\bibitem{CLEM13}
S.~Cl\'emen\c{c}on and J.~Jakubowicz.
\newblock {Scoring anomalies: a M-estimation approach}.
\newblock 2013.

\bibitem{CLEM14}
S.~Cl\'emen\c{c}on and S.~Robbiano.
\newblock Anomaly ranking as supervised bipartite ranking.
\newblock In {\em Proceedings of ICML 2014}, 2014.

\bibitem{hartigan1987estimation}
J.A. Hartigan.
\newblock Estimation of a convex density contour in two dimensions.
\newblock {\em Journal of the American Statistical Association},
  82(397):267--270, 1987.

\bibitem{Kolt97}
V.~Koltchinskii.
\newblock M-estimation, convexity and quantiles.
\newblock {\em The Annals of Statistics}, 25(2):435--477, 1997.

\bibitem{Kolt06}
V.~Koltchinskii.
\newblock Local {R}ademacher complexities and oracle inequalities in risk
  minimization (with discussion).
\newblock {\em The Annals of Statistics}, 34:2593--2706, 2006.

\bibitem{muller1991excess}
D.W. M{\"u}ller and G.~Sawitzki.
\newblock Excess mass estimates and tests for multimodality.
\newblock {\em Journal of the American Statistical Association},
  86(415):738--746, 1991.

\bibitem{Polonik95}
W.~Polonik.
\newblock Measuring mass concentrations and estimating density contour
  cluster-an excess mass approach.
\newblock {\em The annals of Statistics}, 23(3):855--881, 1995.

\bibitem{Polonik97}
W.~Polonik.
\newblock Minimum volume sets and generalized quantile processes.
\newblock {\em Stochastic Processes and their Applications}, 69(1):1--24, 1997.

\bibitem{polonik1998}
W.~Polonik.
\newblock The silhouette, concentration functions and ml-density estimation
  under order restrictions.
\newblock {\em The Annals of Statistics}, 26(5):1857--1877, 10 1998.

\bibitem{SPSSW01}
B.~Sch\"olkopf, J.C. Platt, J.~Shawe-Taylor, A.~Smola, and R.~Williamson.
\newblock {Estimating the Support of a High-Dimensional Distribution}.
\newblock {\em Neural Computation}, 13(7):1443--1471, 2001.

\bibitem{ScottNowak06}
C.~Scott and R.~Nowak.
\newblock Learning \uppercase{M}inimum \uppercase{V}olume \uppercase{S}ets.
\newblock {\em Journal of Machine Learning Research}, 7:665--704, 2006.

\bibitem{SHS05}
I.~Steinwart, D.~Hush, and C.~Scovel.
\newblock A classification framework for anomaly detection.
\newblock {\em J. Machine Learning Research}, 6:211--232, 2005.

\bibitem{VertVert}
J.P. Vert and R.~Vert.
\newblock Consistency and convergence rates of one-class svms and related
  algorithms.
\newblock {\em JMLR}, 6:828--835, 2006.

\bibitem{VCTWMS}
K.~Viswanathan, L.~Choudur, V.~Talwar, C.~Wang, G.~Macdonald, and
  W.~Satterfield.
\newblock Ranking anomalies in data centers.
\newblock In R.D.James, editor, {\em Network Operations and System Management},
  pages 79--87. IEEE, 2012.

\end{thebibliography}


\begin{thebibliography}{1}

\bibitem{federer}
H.~Federer.
\newblock {\em Geometric Measure Theory}.
\newblock Springer, 1969.

\end{thebibliography}

\end{document}


%
\runningtitle{Anomaly Ranking and EM-Curves, Supplementary Material}

%

\twocolumn[

\aistatstitle{On Anomaly Ranking and Excess-Mass Curves, Supplementary Material}

\aistatsauthor{ Nicolas Goix \And Anne Sabourin \And Stéphan Clémençon }

\aistatsaddress{ 
UMR LTCI No. 5141 \\
Telecom ParisTech/CNRS \\
Institut Mines-Telecom \\
Paris, 75013, France 
\And UMR LTCI No. 5141 \\
Telecom ParisTech/CNRS \\
Institut Mines-Telecom \\
Paris, 75013, France 
\And UMR LTCI No. 5141 \\
Telecom ParisTech/CNRS \\
Institut Mines-Telecom \\
Paris, 75013, France } 
]

\section{Illustrations}

\begin{figure}[!h!]
\centering
\includegraphics[width=\linewidth,height=7.5cm]{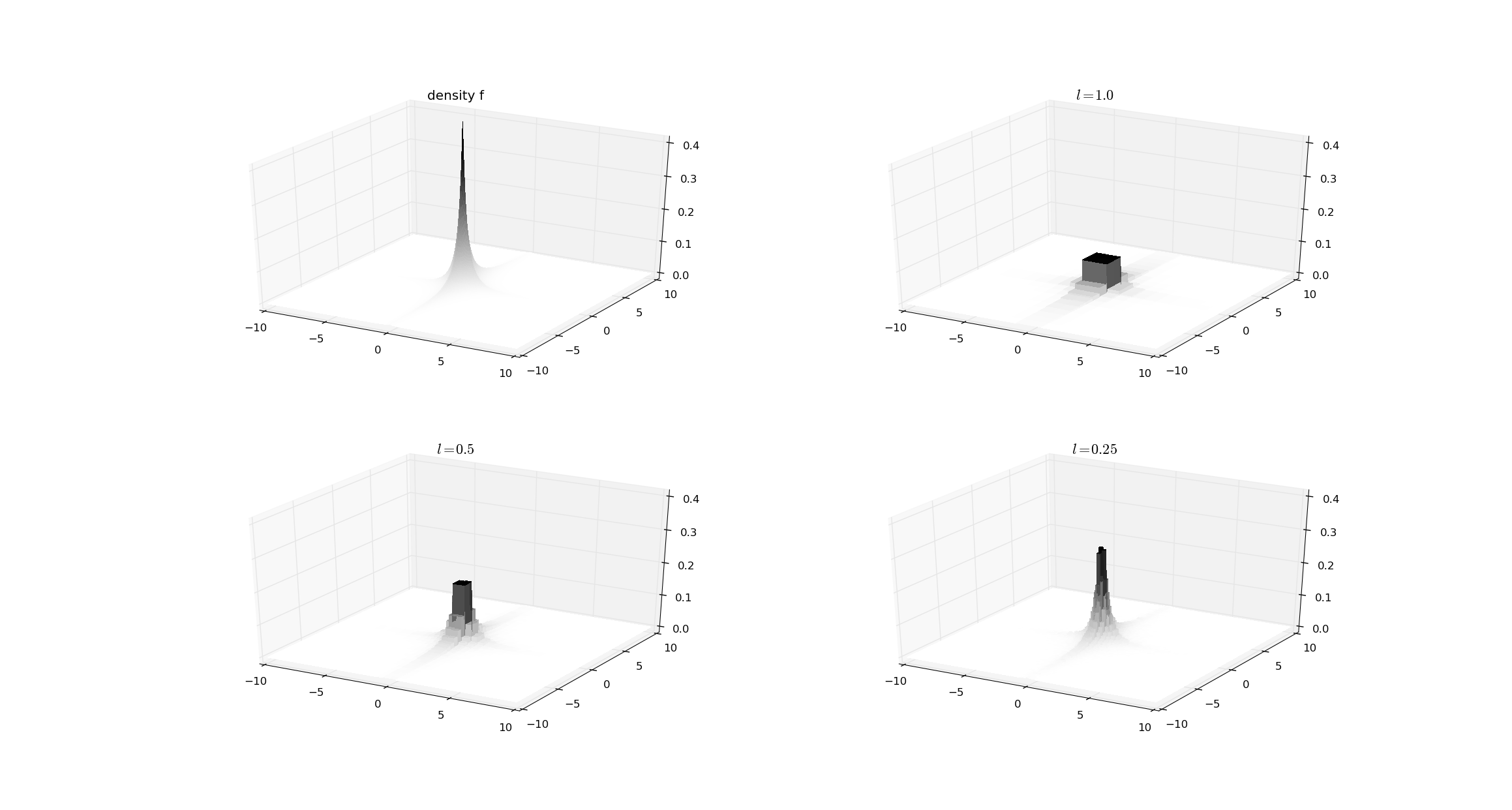}
\caption{density and scoring functions} 
\label{scoring3D}
\end{figure}

Note that the scoring function we built in Algorithm \ref{algo1} is an estimator of the density $f$ (usually called the silhouette), since $f(x)=\int_{0}^\infty \mathds{1}_{f \ge t}dt=\int_{0}^\infty \mathds{1}_{\Omega^*_t}dt$ and $s(x):= \sum_{k=1}^K (t_k-t_{k-1}) \mathds{1}_{x \in \hat{\Omega}_{t_k} }$ which is a discretization of $\int_{0}^\infty \mathds{1}_{\hat \Omega_t}dt$. This fact is illustrated in Fig. \ref{scoring3D}

\section{Detailed Proofs}
\subsection*{Proof of Proposition \ref{derive}} Let $t>0$. Recall that $EM^{*}(t)=\alpha(t)-t \lambda(t)$ where $\alpha(t)$ denote the mass at level $t$, namely $\alpha(t)=\mathbb{P}(f(X) \ge t)$, and $\lambda(t)$ denote the volume at level $t$, i.e. $\lambda(t)=Leb(\{x, f(x) \ge t\})$. For $h>0$, let $A(h)$ denote the quantity $A(h)=1/h(\alpha(t+h)-\alpha(t))$ and $B(h)=1/h (\lambda(t+h)-\lambda(t))$. It is straightforward to see that $A(h)$ and $B(h)$ converge when $h \rightarrow 0$, and expressing $EM^{*'}=\alpha'(t)-t\lambda'(t)-\lambda(t)$, it suffices to show that $\alpha'(t)-t\lambda'(t)=0$, namely $\lim_{h \rightarrow 0} A(h) - t~B(h) = 0$. Now we have $A(h)-t~B(h)~=~\frac{1}{h} \int_{t \le f \le t+h}f-t ~\le~ \frac{1}{h} \int_{t \le f \le t+h} h  ~=~ Leb(t \le f \le t+h) \rightarrow 0$ because $f$ has no flat part.

\subsection*{Proof of Lemma \ref{evident}:}

On the one hand, for every $\Omega$ measurable, 
\begin{align*}
\mathbb{P}(X \in \Omega)-t~Leb(\Omega)&=\int_\Omega(f(x)-t)dx \\&\le \int_{\Omega \cap \{ f \ge t\}}(f(x)-t)dx \\&\le \int_{\{f \ge t\}}(f(x)-t)dx\\&=\mathbb{P}(f(X) \ge t)-t~Leb(\{ f \ge t\}).
\end{align*}
\noindent
It follows that $\{ f \ge t\} \in \arg\max_{A meas.}\mathbb{P}(X \in A)-t~Leb(A) $.\\

On the other hand, suppose $\Omega \in \arg\max_{A~ meas.}\mathbb{P}(X \in A)-t~Leb(A)$ and $Leb(\{f>t\} \setminus \Omega)>0$. Then there is $\epsilon > 0$ such that $Leb(\{f>t+\epsilon\} \setminus \Omega)>0$ (by sub-additivity of Leb, if it is not the case, then $ Leb(\{f>t\} \setminus \Omega) = Leb(\cup_{\epsilon \in \mathbb{Q}_+}\{ f>t+\epsilon\} \setminus \Omega)=0$ ). We have thus $$\int_{\{f>t\} \setminus \Omega} (f(x)-t)dx > \epsilon.Leb(\{f > t+ \epsilon\} \setminus \Omega) > 0~,$$ so that 
\begin{align*}
\int_{\Omega}(f(x)-t)dx &~\le~ \int_{ \{f>t\} }(f(x)-t)dx \\
&~~~~~~~~~~~~~~~~~- \int_{ \{f>t \} \setminus \Omega}(f(x)-t)dx \\
&~<~ \int_{\{f>t\}}(f(x)-t)dx~,
\end{align*}

 i.e  
\begin{align*}
&\mathbb{P}(X \in \Omega)-t~Leb(\Omega) \\
&~~~~~~~~~~~~~~~<~ \mathbb{P}(f(X) \ge t) - t~Leb(\{x,f(x) \ge t\})   
\end{align*}
\noindent
which is a contradiction: $\{f>t\} \subset \Omega$ Leb-a.s. .\\ 

To show that $ \Omega^*_t \subset \{x, f(x) \ge t\}$, suppose that $Leb(\Omega^*_t \cap \{f<t\}) > 0$. Then by sub-additivity of Leb just as above, there is $\epsilon >0$ s.t $Leb(\Omega^*_t \cap \{f<t-\epsilon\}) > 0$ and $\int_{\Omega^*_t \cap \{f<t-\epsilon\}} f-t \le -\epsilon.Leb(\Omega^*_t \cap \{f<t-\epsilon\})<0$. It follows that $\mathbb{P}(X \in \Omega^*_t)-t~Leb(\Omega^*_t) < \mathbb{P}(X \in \Omega^*_t \setminus \{f<t-\epsilon\})-t~Leb(\Omega^*_t \setminus \{f<t-\epsilon\})$ which is a contradiction with the optimality of $\Omega_t^*$.\\\\

\subsection*{Proof of Proposition \ref{propestim}}

Proving the first assertion is immediate, since $\int_{f \ge t}(f(x)-t)dx \ge \int_{s \ge t}(f(x)-t)dx$.
Let us now turn to the second assertion. We have:
\begin{align*}
EM^*(t)-EM_s(t)&=\int_{f>t}(f(x)-t)dx\\&~~~~~~~~~~~~~-~\sup_{u>0}\int_{s>u}(f(x)-t)dx\\
&=\inf_{u>0} \int_{f>t}(f(x)-t)dx     \\&~~~~~~~~~~~~~~~-~\int_{s>u}(f(x)-t)dx~,
\end{align*}
yet:
\begin{align*}
&\int_{\{f>t\}\setminus\{s>u\}}(f(x)-t)dx + \int_{\{s>u\}\setminus\{f>t\}}(t-f(x))dx\\
&~~~\le (\|f\|_\infty-t).Leb\Big(\{f>t\} \setminus \{s>u\}\Big)\\&~~~~~~~~~~~~~~~~~~~~~~~~~~~~~~~~+~ t~Leb\Big(\{s>u\} \setminus \{f>t\}\Big),
\end{align*}
 so we obtain:
\begin{align*}
EM^*(t)-EM_s(t)  &\le~ \max(t,\|f\|_\infty-t) \\ &~~~~~~~~~~~\times Leb\Big(\{s>u\} \Delta \{f>t\}\Big) \\
&\le~ \|f\|_\infty .Leb\Big(\{s>u\} \Delta \{f>t\}\Big) .
\end{align*}

\noindent To prove the third point, note that:

\begin{align*}
&\inf_{u>0} Leb\Big(\{s>u\} \Delta \{f>t\}\Big)\\&~~~~~~~~~~~~~~~~~~~~~~~~~~~~=~\inf_{T \nearrow} Leb\Big(\{Ts>t\} \Delta \{f>t\}\Big)  
\end{align*}

Yet,
\begin{align*} 
&Leb\Big(\{Ts> t\} \Delta \{f>t\}\Big)\\
&\le{Leb(\{f>t-\|Ts-f\|_\infty\} \smallsetminus \{f>t+\|Ts-f\|_\infty\})}\\
&=\lambda(t-\|Ts-f\|_\infty)~-~\lambda(t+\|Ts-f\|_\infty)\\
&=-\int_{t-\|Ts-f\|_\infty}^{t+\|Ts-f\|_\infty}\lambda'(u)du ~.
\end{align*}
\noindent
On the other hand, we have $\lambda(t)=\int_{\mathbb{R}^d}\mathds{1}_{f(x)\ge t}dx = \int_{\mathbb{R}^d} g(x) \|\nabla f(x)\|dx$ where we let $ g(x) = \frac{1}{\|\nabla f(x)\|} \mathds{1}_{\{x,\|\nabla f(x)\|>0, f(x)\ge t\}}$. The co-area formula (see \cite{federer}, p.249, th3.2.12) gives in this case: $\lambda(t)=\int_{\mathbb{R}} du \int_{f^{-1}(u)}\frac{1}{\|\nabla f(x)\|} \mathds{1}_{\{x,f(x)\ge t\}}d\mu (x) = \int_{t}^\infty du \int_{f^{-1}(u)}\frac{1}{\|\nabla f(x)\|}d\mu (x)$ so that $\lambda'(t)=-\int_{f^{-1}(u)}\frac{1}{\|\nabla f(x)\|}d\mu (x)$.\\

\noindent Let $\eta_\epsilon$ such that $ \forall u > \epsilon,~|\lambda'(u)|= \int_{f^{-1}(u)} \frac{1}{\|\nabla f(x)\|} d\mu(x)<\eta_\epsilon$.
 We obtain:
\begin{align*}
&\sup_{t \in [\epsilon + \inf_{T \nearrow}\|f-Ts\|_\infty,\|f\|_\infty]} EM^*(t)-EM_s(t)\\
&~~~~~~~~~~~~~~~~~~~~~~~~~~~~~~~~ \le 2.\eta_\epsilon.\|f\|_\infty \inf_{T \nearrow}\|f-Ts\|_\infty.
\end{align*}
\noindent In particular, if $\inf_{T \nearrow}\|f-Ts\|_\infty \le \epsilon_1 $, 
 $$\sup_{[\epsilon + \epsilon_1,\|f\|_\infty]}|EM^*-EM_s| \le 2.\eta_{\epsilon}.\|f\|_\infty. \inf_{T \nearrow}\|f-Ts\|_\infty~. $$
\noindent

\subsection*{Proof of Proposition \ref{propmono}}

\noindent Let $i$ in $\{1,...,K\}$. First, note  that:
\begin{align*}
&H_{n,t_{i+1}}(\hat \Omega_{t_{i+1}} \cup \hat \Omega_{t_{i}}) = H_{n,t_{i+1}}(\hat \Omega_{t_{i+1}}) \\ &~~~~~~~~~~~~~~~~~~~~~~~~~~~~~~~~~~~~~+ H_{n,t_{i+1}}(\hat \Omega_{t_{i}} \smallsetminus \hat \Omega_{t_{i+1}}),\\
&H_{n,t_{i}}(\hat \Omega_{t_{i+1}} \cap \hat \Omega_{t_{i}}) = H_{n,t_{i}}(\hat \Omega_{t_{i}}) - H_{n,t_{i}}(\hat \Omega_{t_{i}} \smallsetminus \hat \Omega_{t_{i+1}}).
\end{align*}
It follows that
\begin{align*}
 &H_{n,t_{i+1}}( \hat \Omega_{t_{i+1}} \cup \hat \Omega_{t_{i}}) + H_{n,t_{i}}(\hat \Omega_{t_{i+1}} \cap \hat \Omega_{t_{i}}) 
\\&= H_{n,t_{i+1}}(\hat \Omega_{t_{i+1}}) + H_{n,t_{i}}(\hat \Omega_{t_{i}}) + H_{n,t_{i+1}}(\hat \Omega_{t_{i}} \setminus \hat \Omega_{t_{i+1}}) 
\\&~~~~~~~~~~~~~~~~~~~~~~~~~~~~~~~~~~~~~~~~~- H_{n,t_{i}}(\hat \Omega_{t_{i}} \setminus \hat \Omega_{t_{i+1}})\,,
\end{align*}
with $H_{n,t_{i+1}}(\hat \Omega_{t_{i}} \setminus \hat \Omega_{t_{i+1}}) - H_{n,t_{i}}(\hat \Omega_{t_{i}} \setminus \hat \Omega_{t_{i+1}}) \ge 0$ since $H_{n,t}$ is decreasing in $t$. But on the other hand, by definition of $\hat \Omega_{t_{i+1}}$ and $\hat \Omega_{t_{i}}$ we have:
\begin{align*}
&H_{n,t_{i+1}}(\hat \Omega_{t_{i+1}} \cup \hat \Omega_{t_{i}}) \le H_{n,t_{i+1}}(\hat \Omega_{t_{i+1}})\,,\\
&H_{n,t_{i}}(\hat \Omega_{t_{i+1}} \cap \hat \Omega_{t_{i}}) \le H_{n,t_{i}}(\hat \Omega_{t_{i}})\,.
\end{align*}
\noindent
Finally we get:
\begin{align*}
  &H_{n,t_{i+1}}(\hat \Omega_{t_{i+1}} \cup \hat \Omega_{t_{i}}) = H_{n,t_{i+1}}(\hat \Omega_{t_{i+1}})\,,\\
  &H_{n,t_{i}}(\hat \Omega_{t_{i+1}} \cap \hat \Omega_{t_{i}}) =
  H_{n,t_{i}}(\hat \Omega_{t_{i}})\,.
\end{align*}

\noindent
Proceeding by induction we have, for every $m$ such that $k+m \le K$:
\begin{align*}
&H_{n,t_{i+m}}(\hat \Omega_{t_i} \cup \hat \Omega_{t_{i+1}} \cup ... \cup \hat \Omega_{t_{i+m}}) = H_{n,t_{i+m}}(\hat \Omega_{t_{i+m}})~,\\
&H_{n,t_i}(\hat \Omega_{t_i} \cap \hat \Omega_{t_{i+1}} \cap ... \cap \hat \Omega_{t_{i+m}}) = H_{n,t_i}(\hat \Omega_{t_i})~.\\
\end{align*}
Taking (i=1, m=k-1) for the first equation and  (i=k, m=K-k) for the second completes the proof.\\

\subsection*{Proof of Theorem \ref{compact_support_case}}

We shall use the following lemma:
\begin{lemma}
\label{lemmeMs}
With probability at least $1-\delta$, for $k \in \{1,...,K\}$, $0 \le EM^*(t_{k})-EM_{s_K}(t_k) \le 2 \Phi_n(\delta)$. 
\end{lemma}
\noindent

\textbf{Proof of Lemma \ref{lemmeMs}: } 

Remember that by definition of $\hat \Omega_{t_k}$: $H_{n,t_k}(\hat \Omega_{t_k})=\max_{\Omega \in \mathcal{G}} H_{n,t_k}(\Omega) $ and note that:
$$EM^*(t_k)=\max_{\Omega~ meas.} H_{t_k}(\Omega)=\max_{\Omega \in \mathcal{G}} H_{t_k}(\Omega) \ge H_{t_k}(\hat \Omega_{t_k}). $$

On the other hand, using (\ref{penality}),  with probability at least $1-\delta$, for every $G \in \mathcal{G},~ |\mathbb{P}(G)-\mathbb{P}_n (G)| \leq \Phi_n(\delta)$. 
Hence, with probability at least $1-\delta$, for all $\Omega \in \mathcal{G}$ :
\begin{align*}
H_{n,t_k}(\Omega)-\Phi_n(\delta) \le H_{t_k}(\Omega) \le H_{n,t_k}(\Omega) + \Phi_n(\delta)
\end{align*}
\noindent so that, with probability at least $(1-\delta)$, for $k \in \{1..,K\}$,
\begin{align*}
&H_{n,t_k}(\hat \Omega_{t_k})-\Phi_n(\delta) \le H_{t_k}(\hat \Omega_{t_k}) 
\\&~~~~~~~~~~~~~~~~~~~~~~~~~ \le EM^*(t_k) 
\\&~~~~~~~~~~~~~~~~~~~~~~~~~\le H_{n,t_k}(\hat \Omega_{t_k}) + \Phi_n(\delta) ~,
\end{align*}
\noindent whereby, with probability at least $(1-\delta)$, for $k \in \{1,..,K\}$,
\begin{align*}
0 \leq EM^*(t_k) - H_{t_k}(\hat \Omega_{t_k}) ~\leq~ 2 \Phi_n(\delta)~.
\end{align*}

The following Lemma is a consequence of the derivative property of $EM^*$ (Proposition~\ref{derive}) 
\begin{lemma}
\label{lemmederive}
Let $k$ in $\{1,...,K-1\}$. Then for every $t$ in $]t_{k+1},t_{k}]$,
$0 \le EM^*(t)-EM^*(t_{k}) \le \lambda(t_{k+1}) (t_{k}-t_{k+1})$ .
\end{lemma}

\noindent Combined with Lemma \ref{lemmeMs} and the fact that $EM_{s_K}$ is non-increasing, and writing $EM^*(t)-EM_{s_K}(t) = (EM^*(t)-EM^*(t_k)) + (EM^*(t_k) - EM_{s_K}(t_k)) + (EM_{s_K}(t_k) - EM_{s_K}(t))$ this result leads to:
\begin{align*}
&\forall k \in~ \{0,...,K-1\},~ \forall t \in~ ]t_{k+1},t_{k}],
\\&0 \le EM^*(t)-EM_{s_K}(t) \le 2 \Phi_n(\delta)+\lambda(t_{k+1})(t_{k}-t_{k+1}) 
\end{align*}
which gives Lemma \ref{theofini} stated in section Technical Details. Notice that we have not yet used the fact that $f$ has a compact support.


The compactness support assumption allows an extension of Lemma \ref{lemmederive} to $k=K$, namely the inequality holds true for $t$ in $]t_{K+1},t_K]=]0,t_K]$ as soon as we let $\lambda(t_{K+1}):=Leb(supp f)$. Indeed the compactness of $supp f$ implies that $\lambda(t) \rightarrow Leb(supp f)$ as $t \rightarrow 0$. Observing that Lemma \ref{lemmeMs} already contains the case $k=K$, this leads to, for $k$ in $\{0,...,K\}$ and $t \in~ ]t_{k+1},t_{k}],~ |EM^*(t)-EM_{s_K}(t)| \le 2 \Phi_n(\delta)+\lambda(t_{k+1})(t_{k}-t_{k+1})$. Therefore, $\lambda$ being a decreasing function bounded by $\lambda(Leb(supp f))$, we obtain the following: with probability at least $1-\delta$, we have for all $t$ in $]0,t_{1}]$:
\begin{align*}
&|\EM^*(t)-\EM_{s_K}(t)| 
\\ &~\le  \left(A+\sqrt{2log(1/\delta)}\right)\frac{1}{\sqrt n}\\&~~~~~~~~~~~~~~~~~~~~~~~ + \lambda(Leb(supp f))\sup_{1\leq k \le K}(t_{k}-t_{k+1}).
\end{align*}


\subsection*{Proof of Theorem \ref{thmprinc}}
\noindent
The first part of this theorem is a consequence of (\ref{fondineq}) combined with:
\begin{align*}
\sup_{t \in ]0,t_N]} &|EM^*(t)-EM_{s_N}(t)| ~\le~ 1-EM_{s_N}(t_N) \\&~~~~~~~~~~~~~~~~~~~~~~~\le~ 1-EM^*(t_N)+2\Phi_n(\delta)~,
\end{align*} 
where we use the fact that $0 \le EM^*(t_N)-EM_{s_N}(t_N) \le 2 \Phi_n(\delta)$ following from Lemma \ref{lemmeMs}.\\
\noindent To see the convergence of $s_N(x)$, note that:

\begin{align*}
&s_N(x)~=~\frac{t_1}{\sqrt n} \sum_{k=1}^{\infty}\frac{1}{(1+\frac{1}{\sqrt n})^k} \mathds{1}_{x \in \hat \Omega_{t_k}} \mathds{1}_{\{k \le N\}}\\&~~~~~~~~~~~~~~~~~~~~~~~~~~~ \le~ \frac{t_1}{\sqrt n} \sum_{k=1}^{\infty} \frac{1}{(1+\frac{1}{\sqrt n})^k} ~<~ \infty,
\end{align*}

and analogically to remark \ref{orderscore} observe that $EM_{s_N} \le EM_{s_\infty}$ so that $\sup_{t \in ]0,t_1]} |EM^*(t)-EM_{s_\infty}(t)| \leq \sup_{t \in ]0,t_1]} |EM^*(t)-EM_{s_N}(t)|$ which prooves the last part of the theorem.

\subsection*{Proof of Lemma \ref{propbiais}}

By definition, for every class of set $\mathcal{H}$, $EM_{\mathcal{H}}^*(t)=\max_{\Omega \in \mathcal{H}}H_t(\Omega)$.
The bias $ EM^*(t)-EM^*_{\mathcal{G}}(t)$ of the model $\mathcal{G}$ is majored by $ EM^*(t)-EM^*_{\mathcal{F}}(t)$ since $\mathcal{F} \subset \mathcal{G}$.
\noindent Remember that $f_{F}(x) := \sum_{i \ge 1} \mathds{1}_{x \in F_i} \frac{1}{|F_i|} \int_{F_i}f(y)dy$ and note that for all $t>0$, $\{ f_{F} > t \} \in \mathcal{F}$. It follows that:
\begin{align*}
&EM^*(t)-EM^*_{\mathcal{F}}(t) = \int_{f>t}(f-t)- \sup_{C \in \mathcal{F} }\int_{C}(f-t) \\
\le& \int_{f>t}(f-t)- \int_{f_{F} > t}(f-t) \mbox{~since~} \{f_{F} > t\} ~\in~ \mathcal{F}  \\
=&\int_{f>t}(f-t)- \int_{f_{F} > t}(f_{F}-t) \\&~~~~~~~~~~~~~~~~~~~~~~~~~~~~~~\mbox{~since~} \forall G \in \mathcal{F}, \int_Gf=\int_Gf_{F}\\
=&\int_{f>t}(f-t)-\int_{f>t}(f_{F}-t)+\int_{f>t}(f_{F}-t)\\&~~~~~~~~~~~~~~~~~~~~~~~~~~~~~~~~~~~~~~~~~~-\int_{f_{F} > t}(f_{F}-t)\\
=&\int_{f>t}(f-f_{F})+\int_{\{f>t\}\setminus \{f_{F}>t\}}(f_{F}-t)\\&~~~~~~~~~~~~~~~~~~~~~~~~~~~~~~~~~~-\int_{\{f_{F}>t\} \setminus \{f>t\}}(f_{F}-t)~.
\end{align*}
\noindent Observe that the second and the third term in the bound are non-positive. Therefore:
\begin{align*}
EM^*(t)-EM^*_{\mathcal{F}}(t) \le \int_{f>t}(f-f_{F}) \le \int_{\mathbb{R}^d}|f-f_{F}|~.
\end{align*}

\subsubsection*{References}
\renewcommand\refname{\vskip -1cm}
\bibliographystyle{plain}
\bibliography{mvset}